\documentclass{article}
\usepackage{ijcai19}

\usepackage{times}
\usepackage{soul}
\usepackage{url}
\usepackage[hidelinks]{hyperref}
\usepackage[utf8]{inputenc}
\usepackage[small]{caption}
\usepackage{graphicx}
\usepackage{amsmath}
\usepackage{booktabs}
\usepackage{algorithm}
\usepackage{algorithmic}
\usepackage{amssymb}

\title{Region Deformer Networks for Unsupervised Depth Estimation from Unconstrained Monocular Videos}

\author{
Haofei Xu$^1$
\and
Jianmin Zheng$^2$\and
Jianfei Cai$^2$\And
Juyong Zhang$^1$\thanks{Corresponding author}
\affiliations
$^1$University of Science and Technology of China \\
$^2$Nanyang Technological University \\
\emails
xhf@mail.ustc.edu.cn,
\{asjmzheng, asjfcai\}@ntu.edu.sg,
juyong@ustc.edu.cn
}

\begin{document}

\maketitle

\begin{abstract}
  While learning based depth estimation from images/videos has achieved substantial progress, there still exist intrinsic limitations. Supervised methods are limited by a small amount of ground truth  or labeled data  and unsupervised methods for monocular videos are mostly based on the static scene assumption, not performing well on real world scenarios with the presence of dynamic objects. In this paper, we propose a new learning based method consisting of DepthNet, PoseNet and Region Deformer Networks (RDN) to estimate depth from unconstrained monocular videos without ground truth supervision. The core contribution lies in RDN for proper handling of rigid and non-rigid motions of various objects such as rigidly moving cars and deformable humans. In particular, a deformation based motion representation is proposed to model individual object motion on 2D images. This representation enables our method to be applicable to diverse unconstrained monocular videos. Our method can not only achieve the state-of-the-art results on standard benchmarks KITTI and Cityscapes, but also show promising results on a crowded pedestrian tracking dataset, which demonstrates the effectiveness of the deformation based motion representation. Code and trained models are available at \url{https://github.com/haofeixu/rdn4depth}.
\end{abstract}

\section{Introduction}

Depth sensing plays an important role in 3D scene understanding. For instance, it is crucial for robots to be aware of how far the surrounding objects are away from themselves, which helps robots keep clear of obstacles and adjust future behaviour. Recently, learning based single-image depth estimation has attracted a lot of attention due to the rapid progress of Convolutional Neural Networks (CNN). Supervised methods \cite{eigen2014depth,li2015depth,liu2016learning} aim at learning a mapping from color image to per-pixel depth by neural networks. However, these methods require a large quantity of color-depth pairs and collecting such dataset is challenging, especially in outdoor scenarios. 
Unsupervised learning gets rid of the dependence on ground truth depth and shows a promising direction.  The key idea of unsupervised learning is to use the warping based image reconstruction loss between adjacent frames to guide the learning process. Several methods have been proposed to use stereo images to estimate depth \cite{garg2016unsupervised,godard2017unsupervised}. Although no ground truth depth is required, the stereo images are still not as common as monocular videos and they need to be carefully synchronized.

\begin{figure}
    \centering
    \includegraphics[width=\linewidth]{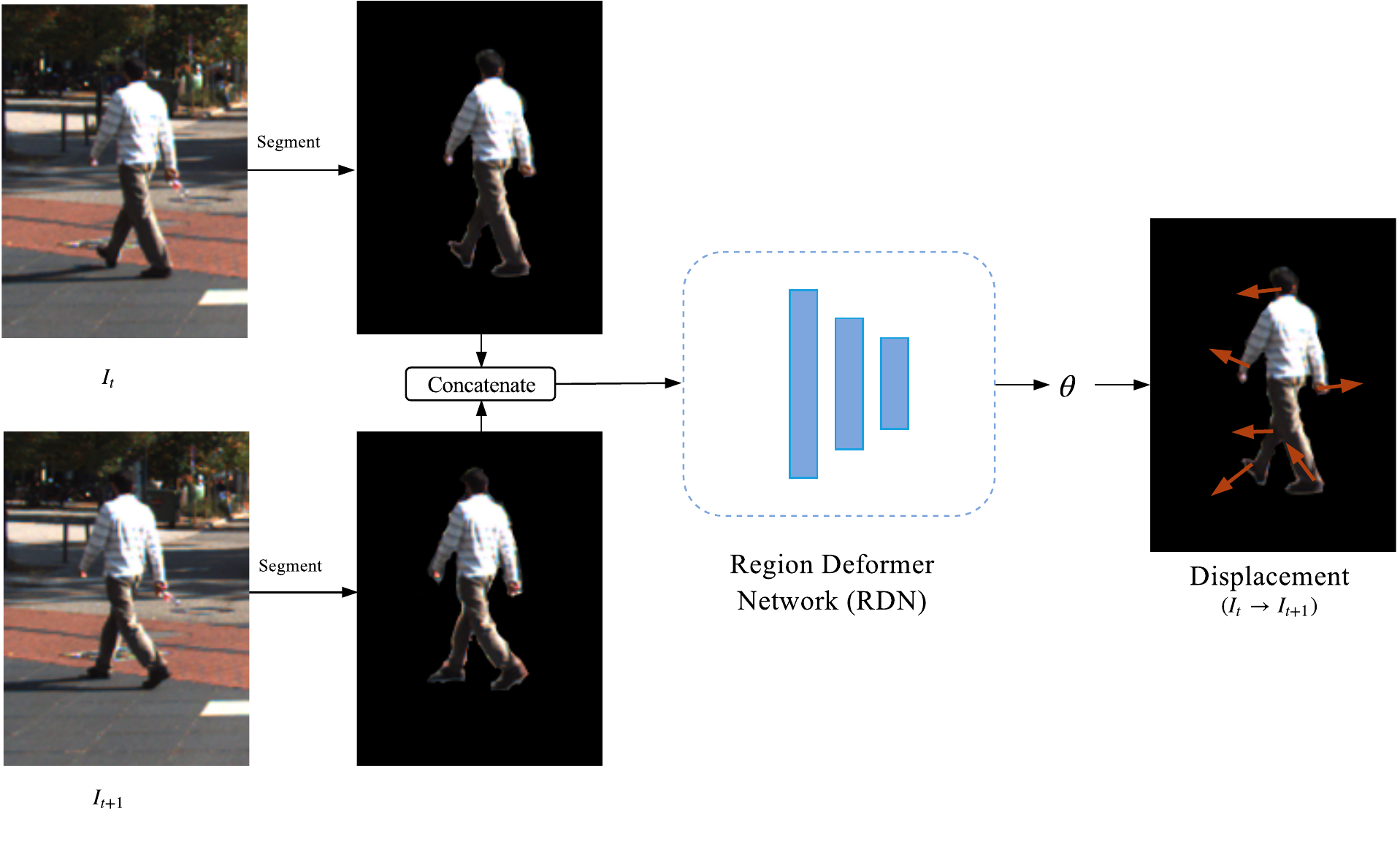}
    \caption{Illustration of our proposed Region Deformer Network (RDN), which is designed to learn the motion of each object between adjacent frames. The input is the concatenation of the same object segmented from adjacent frames $I_t$ and $I_{t+1}$. The object motion is modeled by the bicubic function $f_{\theta}: \mathbb{R}^2 \to \mathbb{R}^2$, which maps pixel location $(x, y)$ to the displacement $(\Delta x, \Delta y)$ on the adjacent frame. The outputs of RDN are the parameters $\theta$ of $f$, which are learned without any predefined correspondences (see Sec.~\ref{sec:rdn} for detail).}
    \label{fig:rdn}
\end{figure}

In this paper, we consider unsupervised depth estimation from monocular videos. In practice, videos are ubiquitous and unlimited. Many previous works in this line are based on the static scene assumption \cite{mahjourian2018unsupervised,wang2018learning,zhou2017unsupervised}, where only camera motion is considered, leading to inaccurate results for moving objects. Several works tried to explicitly model object motion, either with optical flow \cite{yin2018geonet} or SE(3) transforms \cite{casser2019struct2depth} by assuming rigid motion of objects. However, \citeauthor{yin2018geonet}~\shortcite{yin2018geonet} do not report obvious improvement even with the residual flow learning scheme and \citeauthor{casser2019struct2depth}~\shortcite{casser2019struct2depth} only model the motion of rigid objects like cars in driving scenes. For real world scenarios, deformable objects are often present in various forms such as pedestrians and animals. 


We observe that the coupling of camera motion and individual object motion 
often causes ambiguities and may confuse the learning process in dynamic scenes. Therefore we propose to disentangle camera motion and individual object motion between adjacent frames by introducing an additional transformation $f_{\theta}: \mathbb{R}^2 \to \mathbb{R}^2$ for every independently moving object to deform itself to adjacent frame, as illustrated in Fig.~\ref{fig:rdn}. The transformation is modeled as a bicubic deformation and the transformation parameters $\theta$ are learned by a CNN, which is fully guided by image appearance dissimilarity (see Sec.~\ref{formulation} for detail). To realize the individual transformation, the existing instance segmentation method, Mask R-CNN \cite{he2017mask}, is used to segment objects out in each frame, which is only needed at training time and helps the motion representation to learn a better depth network.

The paper has made the following contributions:
\begin{itemize}
    \item We present a learning based method to estimate depth from unconstrained monocular videos. The method consists of DepthNet, PoseNet and Region Deformer Networks, which does not need ground truth supervision.
    \item We propose a deformation based motion representation to model the non-rigid motion of individual objects between adjacent frames on 2D images. This representation is general and enables our method to be applicable to unconstrained monocular videos.
    \item We conduct extensive experiments on three datasets across diverse scenes. Our method can not only achieve the state-of-the-art performance on standard benchmarks KITTI and Cityscapes, but also show promising results on a crowded pedestrian tracking dataset, which validates the effectiveness of the proposed deformation based motion representation.
\end{itemize}

\section{Related Works}
This section briefly reviews some learning based depth estimation work that is most related to ours. 

\textbf{Supervised Depth Estimation.}
Many supervised methods have been developed to estimate depth~\cite{eigen2014depth,li2015depth,liu2016learning,xie2016deep3d}. These methods use CNN to learn a mapping from RGB images to depth maps. However, they need a dataset with ground truth depth which is hard to acquire, especially in outdoor scenarios, and hence limits its applicability.
Several works try to resolve this limitation by using synthetic data \cite{mayer2018makes,zheng2018t2net} or images from Internet \cite{chen2016single,li2018megadepth}, but special care must be taken to generate high quality data, which can be very time-consuming.

{\bf Unsupervised Depth Estimation.}
Unsupervised approaches use image reconstruction loss between adjacent frames to provide self-supervision.
\citeauthor{garg2016unsupervised}~\shortcite{garg2016unsupervised} propose to use calibrated stereo pairs as supervision to train a single view depth CNN. \citeauthor{godard2017unsupervised}~\shortcite{godard2017unsupervised}
 further improve the performance by imposing left-right consistency constraints. 
\citeauthor{zhou2017unsupervised}~\shortcite{zhou2017unsupervised} propose to learn depth and ego-motion from monocular videos under the static scene assumption, with an additional learned explainability mask to ignore the motion of objects. \citeauthor{yin2018geonet}~\shortcite{yin2018geonet} propose to learn a residual flow to handle the motion of objects. \citeauthor{zou2018df}~\shortcite{zou2018df} jointly learn depth and optical flow from monocular videos with a cross-task consistency loss in the rigid scene.



Our work focuses on depth estimation from monocular videos as videos are more easily available than rectified stereo pairs. The work most similar to ours is \cite{casser2019struct2depth}. However, there are two important differences: (i) \citeauthor{casser2019struct2depth}~\shortcite{casser2019struct2depth} model object motion in 3D with SE(3) transformation, which is good for rigidly moving objects, like cars in driving scenes.  We use a deformation based representation to model object motion in 2D image plane, which is more general to be applicable to diverse real world scenarios. (ii) To handle the common issue that cars moving in front of camera at roughly the same speed are often projected into infinite depth in monocular setting, \citeauthor{casser2019struct2depth}~\shortcite{casser2019struct2depth} propose to impose object size constraints depending on the height of object segmentation mask, which is not suitable for deformable objects as the actual scale can be varied over time. Also, the constraints in \cite{casser2019struct2depth} are learned by a network which can be tricky to find the good hyper-parameters. Instead, we choose to use a simple yet efficient prior inspired from \cite{ranftl2016dense}, which is more general for diverse scenes and has no parameters to learn.

{\bf Learning Geometric Transformation.} Spatial Transformer Networks (STN) \cite{jaderberg2015spatial} build the first learnable module in the network architecture to handle geometry variation of input data, which is realized by learning a global parametric transformation. Deformable ConvNets \cite{dai2017deformable} further extend STN by learning offsets to regular grid sampling locations in the standard convolution. STN and Deformable ConvNets are both aiming at designing network architectures with geometry invariant for supervised tasks like classification and segmentation. Our deformation based motion representation aims at learning a transformation for each of individual objects to model object motion between adjacent frames. 

\section{Proposed Method}
\label{sec:method}
\begin{figure}[!t]
\centering
\includegraphics[width=\linewidth]{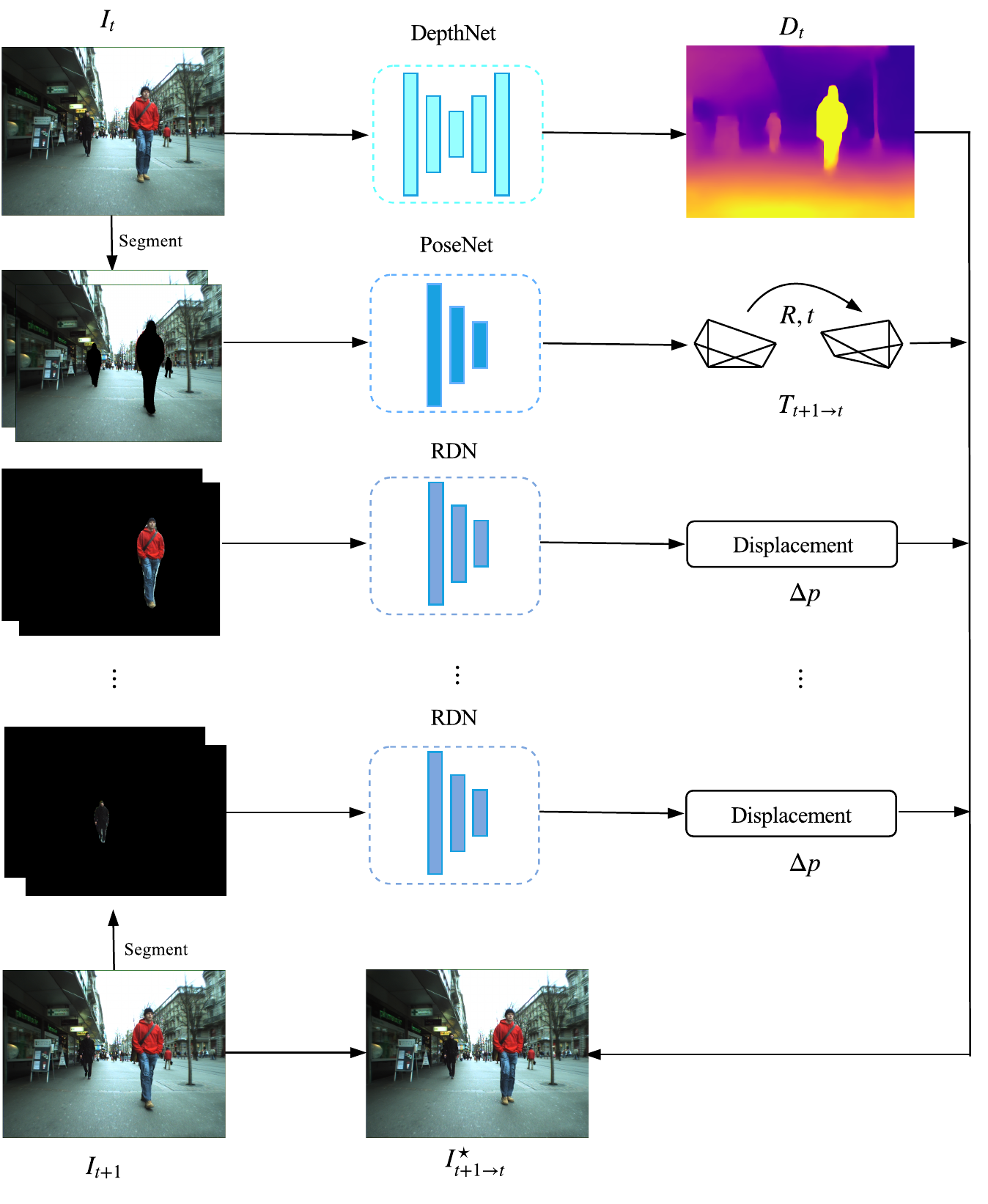}
\caption{Overview of our proposed framework. DepthNet and PoseNet predict $I_t$'s depth $D_t$ and the camera pose transformation $T_{t+1 \to t}$ between $I_t$ and $I_{t+1}$, respectively. 
The motion displacement $\Delta p$ of each object is predicted by a Region Deformer Network (RDN). With $D_t, T_{t+1\to t}, \Delta p$ and $I_{t+1}$, we reconstruct a synthetic image $I^{\star}_{t+1 \to t}$ corresponding to $I_t$, the appearance dissimilarity between $I^\star_{t+1 \to t}$ and $I_t$ provides training signal of our framework.}
\label{fig:overview}
\end{figure}

This section presents our generic framework for unsupervised depth estimation from unconstrained monocular videos. 
The input is a sequence of video frames $\{I_t \}_{t=1}^N, I_t \in \mathbb{R}^{H \times W \times C}$, where $H, W, C$ represent frame $I_t$'s height, width and number of channels, respectively.  Our goal is to estimate the corresponding depth maps $\{ D_t \}_{t=1}^N, D_t \in \mathbb{R}^{H \times W}$. For this purpose, we build a DepthNet to learn the mapping from color image to per-pixel depth map, a PoseNet to learn the mapping from two adjacent frames to their relative camera pose transformation, and multiple RDNs in parallel to learn the transformations that model the motion of individual objects from one frame to its adjacent frame.
The overall framework is illustrated in Fig.~\ref{fig:overview}. For two adjacent frames $I_t$ and $I_{t+1}$, we first obtain instance segmentation masks from the existing Mask R-CNN model. $I_t$ is fed into the DepthNet to predict its depth $D_t$. The concatenation of $I_t$ and $I_{t+1}$ is fed into the PoseNet to learn the camera pose transformation $T_{t+1 \to t}$ between $I_t$ and $I_{t+1}$, where objects are masked out to avoid motion clue from possibly moving objects. We further use the RDN to model the motion $\Delta p$  of individual objects in parallel. With $D_t, T_{t+1\to t}, \Delta p$ and $I_{t+1}$, we reconstruct a synthetic frame  $I^{\star}_{t+1 \to t}$ corresponding to $I_t$. The appearance dissimilarity between $I^\star_{t+1 \to t}$ and $I_t$ provides training signal of our framework. During testing, only the DepthNet is needed to predict the depth for an input frame.

Below we first give the basic formulation of the loss function, and then describe the deformation based motion representation that explicitly handles object motion on 2D images.

\subsection{Basic Formulation}
\label{formulation}
With $D_t, I_{t+1}, T_{t+1 \to t}$, we synthesize image
\begin{equation}
\label{eq:warp}
\hat{I}_{t+1\to t} = \mathcal{W}(D_t, I_{t+1}, T_{t+1 \to t})
\end{equation}
corresponding to frame $I_t$, where $\mathcal{W}$ is a warping function. We first construct 
$\mathcal{W}$ based on the static scene assumption, and then improve it for dynamic scenes by adding individual object motions in Sec.~\ref{sec:rdn}.

For each pixel coordinates $p = (x, y)$ in frame $I_t$, we can obtain its projected coordinates $\hat{p} = ({\hat{x}, \hat{y}})$ in frame $I_{t+1}$ with the estimated depth $D_t$ and camera transformation $T_{t+1 \to t}$
\begin{equation}
\label{eq:project}
    \hat{p} \sim K T_{t+1 \to t} D_t(p) K^{-1} h(p),
\end{equation}
where $h(p) = (x, y, 1) $ denotes the homogeneous coordinates of $p$ and $K$ denotes the camera intrinsic matrix. We use the bilinear  mechanism used in  \cite{jaderberg2015spatial} to sample frame $I_{t+1}$ to create image $\hat{I}_{t+1 \to t}$.

The appearance dissimilarity between reconstructed image $\hat{I}_{t+1 \to t}$ and $I_t$  is defined as
\begin{equation}
\ell = \rho(I_t, \hat{I}_{t+1 \to t}),
\end{equation}
where function $\rho$ is  a combination of $L_1$ photometric error and Structure Similarity (SSIM) \cite{wang2004image}:
\begin{equation}
\label{eq:dissim}
\rho(I, \hat{I}) = \alpha \frac{1 - \mathrm{SSIM}(I, \hat{I})}{2} + (1 - \alpha) \| I - \hat{I} \|_1.
\end{equation}
To handle occlusion/disocclusion between adjacent frames, per-pixel minimum of the dissimilarities with previous frame and next frame is used as proposed in \cite{godard2018digging}:
\begin{equation}
L_{\mathrm{ap}} = \min(\rho(I_t, \hat{I}_{t+1 \to t}), \rho(I_t, \hat{I}_{t-1 \to t})),
\end{equation}
where $\hat{I}_{t-1 \to t}$ is reconstructed image from $I_{t-1}$.

To regularize the depth, we further impose an image-aware depth smoothness constraint as commonly used in previous works \cite{godard2017unsupervised,yin2018geonet}:
\begin{equation}
    \label{smooth}
    L_{\mathrm{s}} = \sum_{x,y} \| \partial_x D_t \| e^{-\| \partial_x I_t \|} + \| \partial_y D_t \| e^{-\| \partial_y I_t \|}.
\end{equation}

The total loss function is a combination of $L_{\mathrm{ap}}$ and $L_{\mathrm{s}}$ applied on four different image scales, ranging from the model's input resolution, to an image that is $1/8$ in height and width: 
\begin{equation}
\label{eq:baseline}
L = \sum_{k=0}^3 L_{\mathrm{ap}}^{(k)} + \lambda \cdot \frac{1}{2^k} L_{\mathrm{s}}^{(k)},
\end{equation}
where $\lambda$ is a hyper-parameter. 
We use Eq.~\ref{eq:baseline} as our baseline model, which is similar to that in \cite{casser2019struct2depth}.

\subsection{Region Deformer Networks}
\label{sec:rdn}

For dynamic scenes, individual objects may have their own motions  besides camera motion. We propose to explicitly model individual object motion on 2D images. Specifically, our goal is to learn the displacement vector $\Delta p = (\Delta x, \Delta y)$ of every pixel belonging to moving objects.  Then the corresponding pixel of $p$ can be found by $p^\star = \hat{p} + \Delta p$. This process is accomplished by learning a function $f: \mathbb{R}^2 \to \mathbb{R}^2$ for each object to map $(x, y)$ to the displacement $(\Delta x, \Delta y)$. The basic requirement for $f$ is that it should be flexible enough to model non-rigid object motion. We choose the bicubic function, which is expressed as 
\begin{equation}
\label{bicubic}
\begin{cases}
\Delta x = \displaystyle\sum_{i=0}^3 \sum_{j=0}^3 a_{ij} x^i y^j, \\
\Delta y = \displaystyle\sum_{i=0}^3 \sum_{j=0}^3 b_{ij} x^i y^j.
\end{cases}
\end{equation}
The bicubic function is widely used in various applications. It has low computational cost, and meanwhile it contains 32 coefficients, providing sufficient degrees of freedom for modeling freeform deformation (or motion).
 
To learn the transformation parameters $\{a_{ij}\}_{i,j=0}^3$ and $\{b_{ij}\}_{i,j=0}^3$, we design the Region Deformer Networks (RDN).
The workflow of the RDN is illustrated in Fig.~\ref{fig:rdn}. Given two adjacent frames $I_t$ and $I_{t+1}$, an instance segmentation model \cite{he2017mask} is used to segment objects within each frame. Let ${M_t^i}$ and ${M_{t+1}^i}$ denote the $i$-th binary object segmentation masks in frame $I_t$ and $I_{t+1}$, respectively. We first compute the reconstructed image $\hat{I}_{t+1 \to t}$ and mask $\hat{M}_{t+1 \to t}^i$ by camera motion using Eq.~\ref{eq:warp}, which eliminates camera motion between adjacent frames. The input of RDN is the concatenation of objects $O_t^i = I_t \odot M_t^i$ and $\hat{O}_{t+1 \to t}^i = \hat{I}_{t+1 \to t} \odot \hat{M}_{t+1 \to t}^i$ in $I_t$ and $\hat{I}_{t+1 \to t}$, respectively, where $\odot$ denotes element-wise multiplication, making only the $i$-th object visible for every independent RDN. When multiple objects exist in a single frame, multiple RDNs are used in parallel to model every independent object motion. The outputs of RDN are the parameters of the transformation. By applying the transformation to its corresponding object, we obtain the displacement of the pixels belonging to that object.

Now we are ready to refine the function $\mathcal{W}$ by defining a new warping function $\mathcal{W}^\star$. If a pixel $p$ belongs to static background, we compute its correspondence by Eq.~\ref{eq:project}. If $p$ belongs to moving objects, its correspondence is found by $p^\star = \hat{p} + \Delta p$, where $\hat{p}$ is from camera motion using Eq.~\ref{eq:project} and $\Delta p$ is from object motion obtained by RDN. In general, for every pixel $p$ in frame $I_t$, we can get its correspondence by
\begin{equation}
    \label{eq:deform}
    p^\star = \hat{p} + M_t(p) \cdot \Delta p,
\end{equation}
where  $M_t = \sum_{i=1}^S M_t^i$ is the binary instance segmentation mask for $I_t$: $M_t(p)$ is 1 if $p$ belongs to moving objects; otherwise 0, and $S$ is the total number of objects in $I_t$. By modeling individual object motion with RDN, we can get more accurately reconstructed image
\begin{equation}
\label{eq:new}
I^\star_{t+1 \to t} = \mathcal{W}^\star(D_t, I_{t+1}, T_{t+1 \to t}, f_1, f_2, \cdots, f_S),
\end{equation}
where $\{f_i\}_{i=1}^S$ are the individual transformations learned by the RDN. Similarly, we can generate image $I^\star_{t-1 \to t}$.

\subsection{Object Depth Prior}
\label{sec:prior}
Depth estimation from monocular videos has an issue that objects moving with camera at the roughly same speed are often projected to infinite depth, as this shows very little appearance change, resulting in low reprojection error \cite{godard2018digging}. \citeauthor{casser2019struct2depth}~\shortcite{casser2019struct2depth} propose to impose object size constraints by additionally learning the actual scales of objects. These constraints are internally based on the assumption that object scales are fixed. However, in real world scenarios, deformable objects are often present, like pedestrians and animals, which are not applicable for these constrains. Furthermore, as the actual scales are also learned during the training process, it can be tricky to find the good hyper-parameters. 

We propose to use a simple yet efficient prior inspired from \cite{ranftl2016dense}: objects are supported by their surrounding environment, which is often true in most real world scenes. This prior can be used by requiring the depths of moving objects to be smaller or equal to their horizontal neighbors. However, noticing that overlapping objects may exist in the real world, which might violate this depth prior, we thus introduce a soft constraint, which is formulated as
\begin{equation}
    \label{eq:prior}
    L_{\mathrm{prior}} = \max(d_{\mathrm{obj}} - d_{\mathrm{neigh}} - \delta, 0),
\end{equation}
where $d_{\mathrm{obj}}$ is the mean depth of an individual object, $d_{\mathrm{neigh}}$ is the mean depth of its horizontal neighbors in a small range, and $\delta$ is a small positive number to handle exceptions violating our depth prior. 
The main idea here is to use the depth prior to prevent the degenerated cases of infinite depth. Note that if this prior is satisfied, which happens most of the time, Eq. \ref{eq:prior} actually has no use (i.e., the loss becomes $0$). 

By incorporating the  RDN and object depth prior, our final loss function can be expressed as
\begin{equation}
    \label{eq:final}
    L^\star = \sum_{k=0}^3 L_{\mathrm{ap}}^{\star(k)} + \lambda \cdot \frac{1}{2^k} L_s^{(k)} + \mu \cdot  L^{(k)}_{\mathrm{prior}},
\end{equation}
where $\mu$ is a hyper-parameter. The only difference between  $L_{\mathrm{ap}}^{(k)}$ and $L_{\mathrm{ap}}^{\star(k)}$ is that we replace $I_{t+1 \to t}, I_{t-1 \to t}$ in Eq.~\ref{eq:warp} with $I^\star_{t+1 \to t}, I^\star_{t-1 \to t}$. We treat Eq.~\ref{eq:final} as our motion model.

\section{Experiments}

\subsection{Datasets}

We conduct experiments on several datasets across diverse scenes, including not only standard benchmarks KITTI and Cityscapes, but also a publicly available pedestrian tracking dataset. The details of  each dataset are given below.

{\bf KITTI}. The KITTI dataset \cite{geiger2012we} is a popular benchmark for scene understanding in outdoor driving scenario. Only the monocular video sequences are used for training and no ground truth depth is needed. 
The performance is evaluated on Eigen split \cite{eigen2014depth} using the standard evaluation protocol.

{\bf Cityscapes}. The Cityscapes \cite{cordts2016cityscapes} is an outdoor driving dataset similar to KITTI, but with more moving objects. 
We use this dataset for training and the evaluation is done on Eigen split.

{\bf Pedestrian Tracking Dataset}.  To validate that our motion representation is general enough to model deformable objects, we collect videos from a publicly available pedestrian tracking dataset \cite{ess2009robust}, which was recorded on a crowded pedestrian zone. This dataset is very challenging as large human deformations are frequently observed.

\begin{figure*}[t]
\centering
\setlength{\tabcolsep}{2pt}

\begin{tabular}{lcccc}

\rotatebox{90}{{\scriptsize  \quad Input}} &
\includegraphics[width=0.23\linewidth]{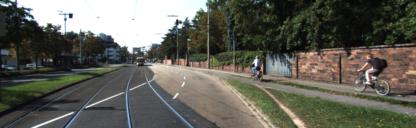} &
\includegraphics[width=0.23\linewidth]{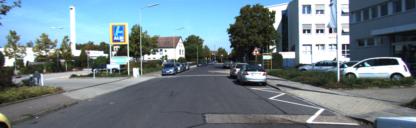} &
\includegraphics[width=0.23\linewidth]{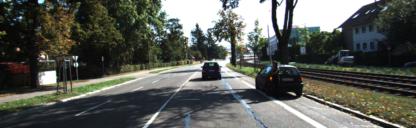} &
\includegraphics[width=0.23\linewidth]{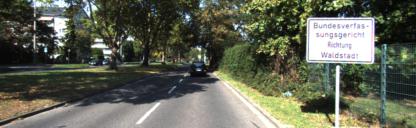} \\

\rotatebox{90}{{\scriptsize \citeauthor{zhou2017unsupervised}}} &
\includegraphics[width=0.23\linewidth]{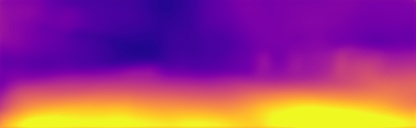} &
\includegraphics[width=0.23\linewidth]{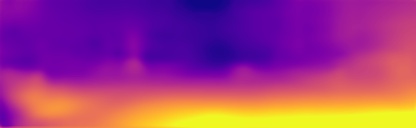} &
\includegraphics[width=0.23\linewidth]{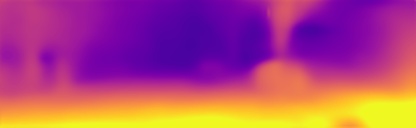} &
\includegraphics[width=0.23\linewidth]{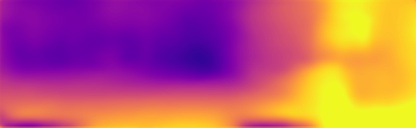} \\

\rotatebox{90}{{\scriptsize \citeauthor{yin2018geonet}}} &
\includegraphics[width=0.23\linewidth]{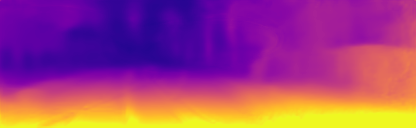} &
\includegraphics[width=0.23\linewidth]{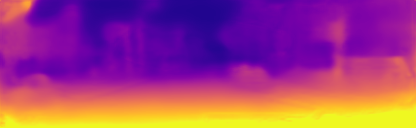} &
\includegraphics[width=0.23\linewidth]{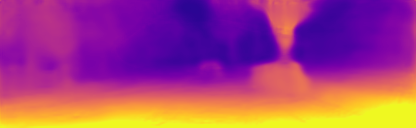} &
\includegraphics[width=0.23\linewidth]{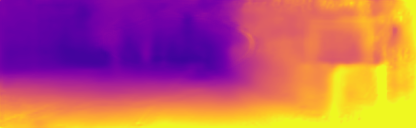} \\

\rotatebox{90}{{\scriptsize \citeauthor{zou2018df}}} &
\includegraphics[width=0.23\linewidth]{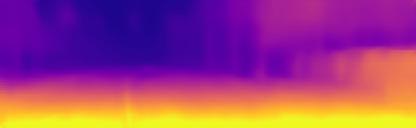} &
\includegraphics[width=0.23\linewidth]{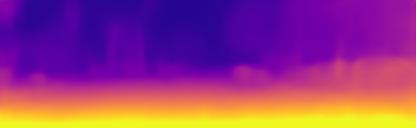} &
\includegraphics[width=0.23\linewidth]{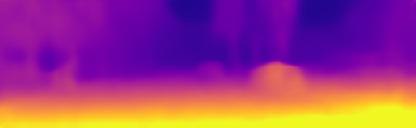} &
\includegraphics[width=0.23\linewidth]{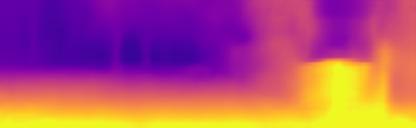} \\

\rotatebox{90}{{\scriptsize \citeauthor{casser2019struct2depth}}} &
\includegraphics[width=0.23\linewidth]{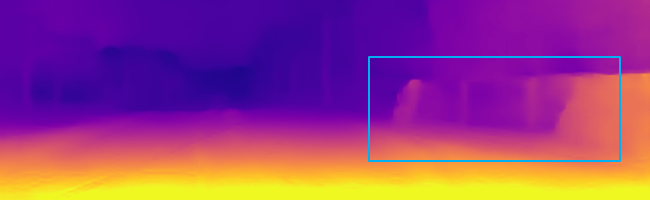} &
\includegraphics[width=0.23\linewidth]{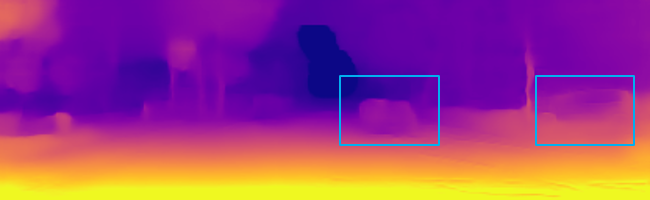} &
\includegraphics[width=0.23\linewidth]{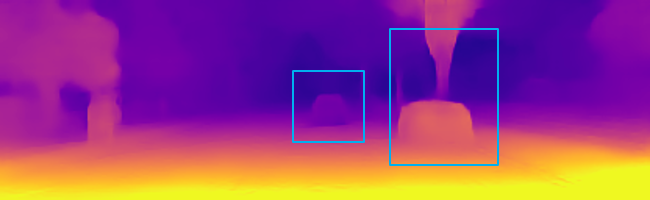} &
\includegraphics[width=0.23\linewidth]{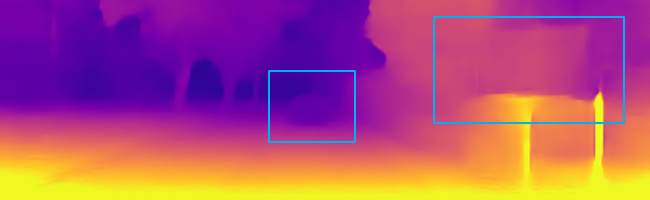} \\

\rotatebox{90}{{\scriptsize  \quad Ours}} &
\includegraphics[width=0.23\linewidth]{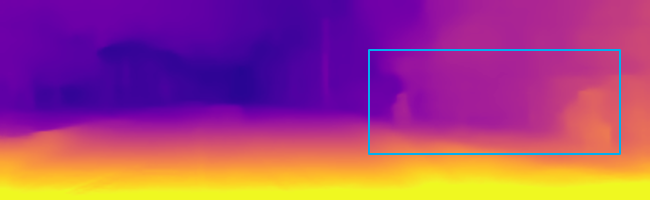} &
\includegraphics[width=0.23\linewidth]{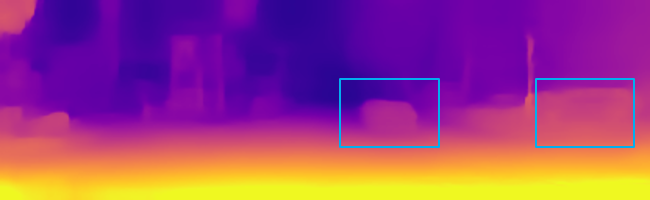} &
\includegraphics[width=0.23\linewidth]{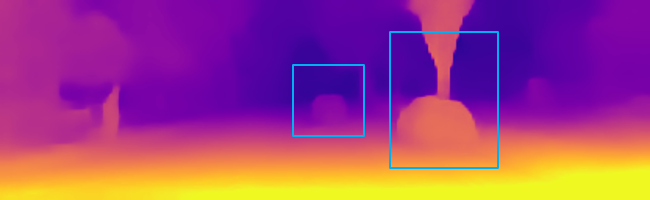} &
\includegraphics[width=0.23\linewidth]{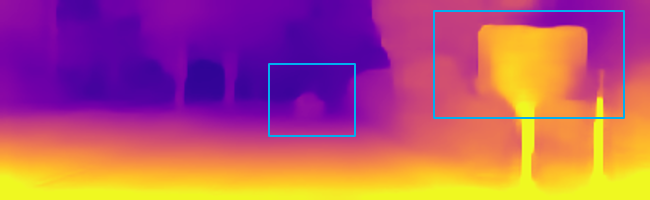} \\

\rotatebox{90}{{\scriptsize \hspace{0.1cm} GT}} &
\includegraphics[width=0.23\linewidth]{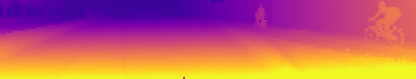} &
\includegraphics[width=0.23\linewidth]{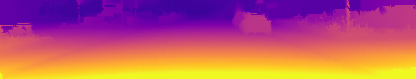} &
\includegraphics[width=0.23\linewidth]{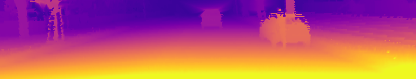} &
\includegraphics[width=0.23\linewidth]{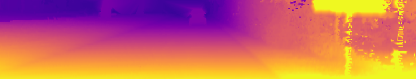} \\

\end{tabular}

\caption{Qualitative comparisons on Eigen test split. GT denotes ground truth depth, which is interpolated for visualization purpose, and the upper regions are cropped as they are not available. Compared with other algorithms, our method can better capture scene structures and moving objects like cars and riding people.}
\label{fig:qualitative}
\end{figure*}

\begin{table*}[!h]
\centering


\begin{tabular*}{\textwidth}{c @{\extracolsep{\fill}} |c|cccc|cccc}
\hline
Method & Dataset & Abs Rel & Sq Rel & RMSE & RMSE log &
$\delta < 1.25$ & $\delta < 1.25^2$ & $\delta < 1.25^3$ \\
\hline
\citeauthor{zhou2017unsupervised} & K & 0.208 & 1.768 & 6.856 & 0.283 & 0.678 & 0.885 & 0.957 \\
\citeauthor{mahjourian2018unsupervised} & K & 0.163 & 1.240 & 6.220 & 0.250 & 0.762 & 0.916 & 0.968 \\
\citeauthor{yin2018geonet} & K & 0.155 & 1.296 & 5.857 & 0.233 & 0.793 & 0.931 & 0.973 \\
\citeauthor{wang2018learning} & K & 0.151 & 1.257 & 5.583 & 0.228 & 0.810 & 0.936 & 0.974 \\
\citeauthor{zou2018df} & K & 0.150 & 1.124 & 5.507 & 0.223 & 0.806 & 0.933 & 0.973 \\
\citeauthor{casser2019struct2depth} & K & 0.1412 & 1.0258 & \bf{5.2905} & \bf{0.2153} & 0.8160 & \bf{0.9452} & \bf{0.9791} \\
Ours (Baseline) & K & 0.1436 & 1.1732 & 5.4468 & 0.2188 & 0.8202 & 0.9428 & 0.9762 \\
Ours (Motion) & K & \bf{0.1377} & \bf{1.0155} & 5.3523 & 0.2165 & \bf{0.8226} & 0.9428 & 0.9762 \\
\hline
\citeauthor{casser2019struct2depth} & C & 0.1876 & 1.3541 & 6.3166 & 0.2641 & 0.7135 & 0.9046 & 0.9667 \\
Ours (Baseline) & C & 0.1929 & 1.6131 & 6.4098 & 0.2680 & 0.7189 & 0.9071 & 0.9641 \\
Ours (Motion) & C & \bf{0.1816} & \bf{1.3160} & \bf{6.1484} & \bf{0.2573} & \bf{0.7263} & \bf{0.9124} & \bf{0.9677} \\
\hline

\end{tabular*}

\caption{Unsupervised monocular depth estimation results on Eigen test split of KITTI raw dataset. We use K and C to denote models trained on KITTI and Cityscapes dataset, respectively. Abs Rel, Sq Rel, RMSE and RMSE log are error metrics (lower is better). $\delta < 1.25$, $\delta < 1.25^2$ and $\delta < 1.25^3$ are accuracy metrics (higher is better). The best performance in each group is highlighted in bold. 
}
\label{tab:eigen}
\end{table*}


\begin{table*}[!h]
\centering


\begin{tabular*}{\textwidth}{c @{\extracolsep{\fill}} |c|cccc|cccc}
\hline
Method & Dataset & Abs Rel & Sq Rel & RMSE & RMSE log &
$\delta < 1.25$ & $\delta < 1.25^2$ & $\delta < 1.25^3$ \\
\hline

Ours (w/o prior) & K & 0.1419	 &   1.1635	 &   5.4739	 &   0.2206	 &   0.8217	 &   0.9416	 &   0.9751 \\
Ours (Rigid) & K & 0.1391	&    1.0384	 &   \bf{5.3518}	  &  0.2174	 &   \bf{0.8231}	 &   0.9425 &	    0.9759 \\
Ours (Motion) & K & \bf{0.1377} & \bf{1.0155} & 5.3523 & \bf{0.2165} & 0.8226 & \bf{0.9428} & \bf{0.9762} \\
\hline

Ours (w/o prior) & C & 0.1973	 &   1.9500	  &  6.6228	  &  0.2695	 &   0.7200	  &  0.9074	 &   0.9630
\\
Ours (Rigid) & C & 0.1820	 &   1.3500	 &   6.1657	  &  0.2575	  &  \bf{0.7313}	 &   0.9119	&    0.9672 \\
Ours (Motion) & C & \bf{0.1816} & \bf{1.3160} & \bf{6.1484} & \bf{0.2573} & 0.7263 & \bf{0.9124} & \bf{0.9677} \\
\hline

\end{tabular*}

\caption{Ablation study on Eigen test split of KITTI raw dataset. Ours (w/o prior) is Ours (Motion) without the depth prior, Ours (Rigid) is the model by replacing bicubic transform in Ours (Motion) with rigid motion representation.}
\label{tab:ablation}
\end{table*}

\begin{figure*}
\centering
\setlength{\tabcolsep}{2pt}

\begin{tabular}{lcccc}

\rotatebox{90}{{\scriptsize  \quad Input}} &
\includegraphics[width=0.23\linewidth]{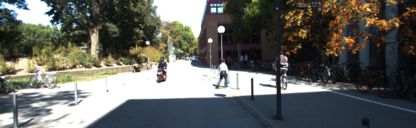} &
\includegraphics[width=0.23\linewidth]{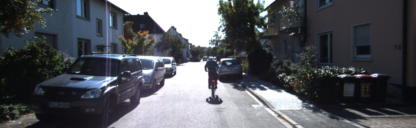} &
\includegraphics[width=0.23\linewidth]{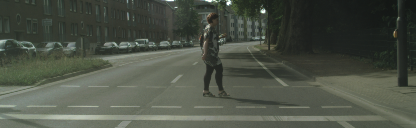} &
\includegraphics[width=0.23\linewidth]{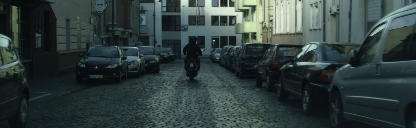} \\

\rotatebox{90}{{\scriptsize \quad Baseline}} &
\includegraphics[width=0.23\linewidth]{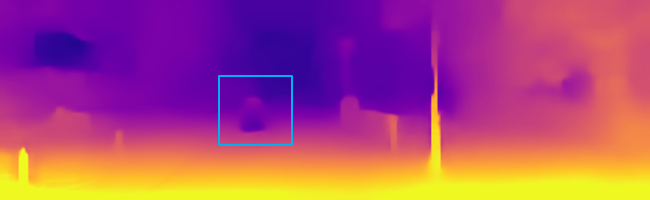} &
\includegraphics[width=0.23\linewidth]{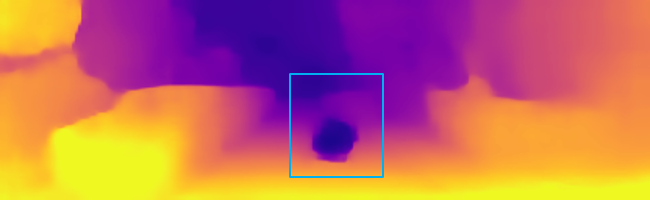} &
\includegraphics[width=0.23\linewidth]{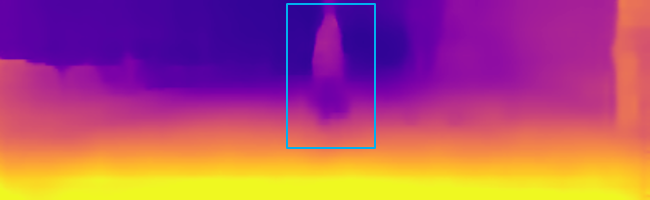} &
\includegraphics[width=0.23\linewidth]{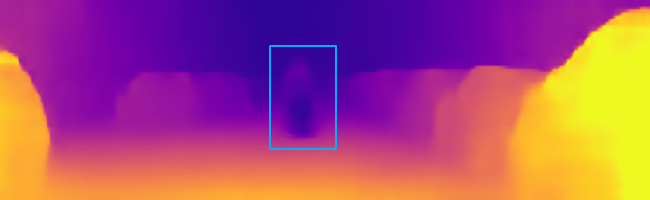} \\

\rotatebox{90}{{\scriptsize \quad Motion}} &
\includegraphics[width=0.23\linewidth]{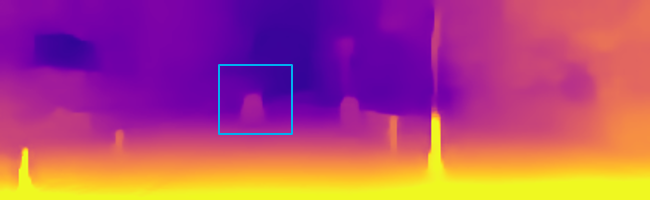} &
\includegraphics[width=0.23\linewidth]{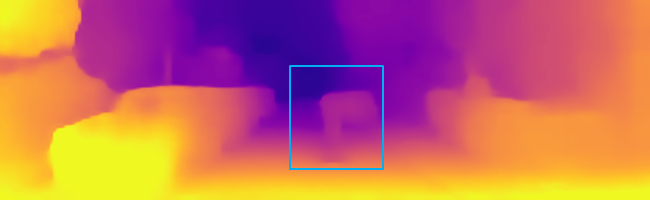} &
\includegraphics[width=0.23\linewidth]{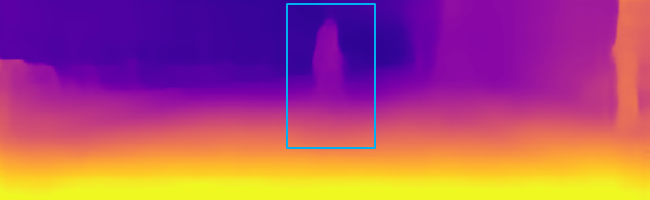} &
\includegraphics[width=0.23\linewidth]{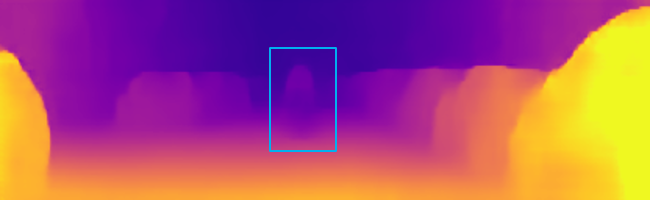} \\

\end{tabular}

\caption{Visualization of depth prediction results on KITTI and Cityscapes datasets. Our motion model is clearly better than the baseline to estimate deformable humans, validating the effectiveness of our deformation based bicubic motion representation.}
\label{fig:compare}
\end{figure*}


\begin{figure}[!h]
\centering
\setlength{\tabcolsep}{2pt}

\begin{tabular}{lccc}

\rotatebox{90}{{\hspace{0.5cm} Input}} &
\includegraphics[width=0.3\linewidth]{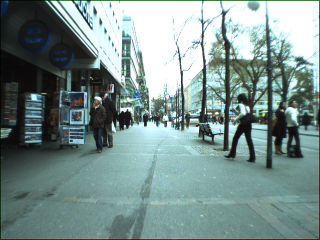} &
\includegraphics[width=0.3\linewidth]{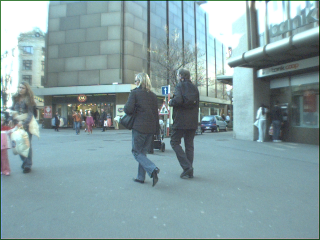} &
\includegraphics[width=0.3\linewidth]{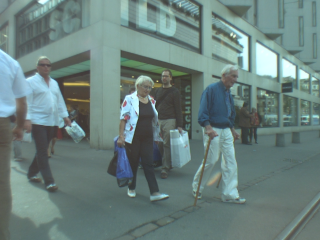} 
\\

\rotatebox{90}{{\hspace{0.35cm} Baseline}} &
\includegraphics[width=0.3\linewidth]{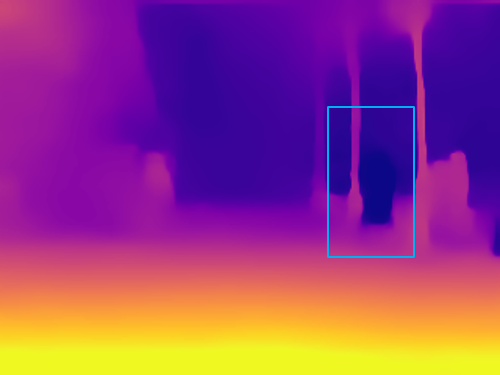} &
\includegraphics[width=0.3\linewidth]{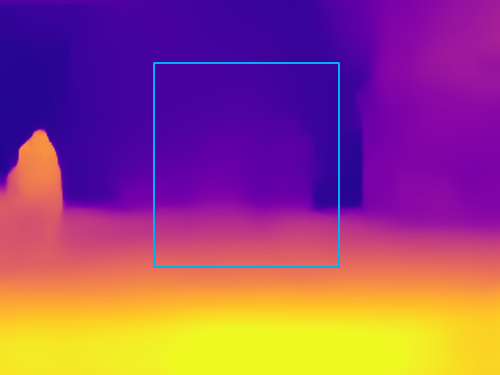} &
\includegraphics[width=0.3\linewidth]{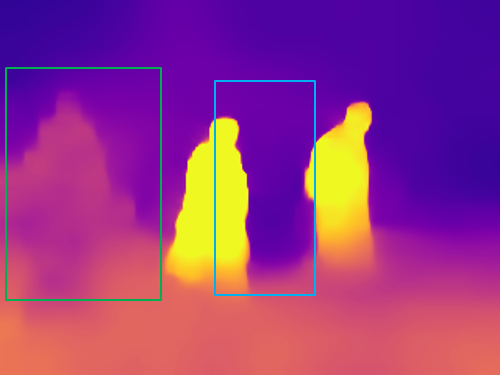} 
\\

\rotatebox{90}{{\hspace{0.45cm} Motion}} &
\includegraphics[width=0.3\linewidth]{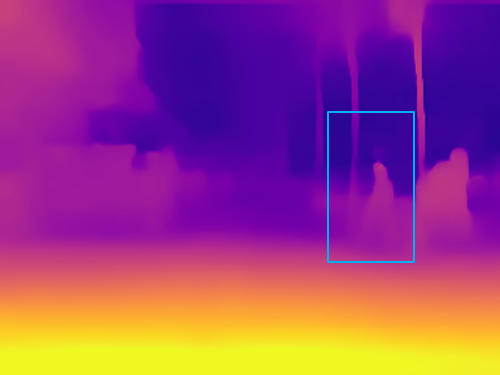} &
\includegraphics[width=0.3\linewidth]{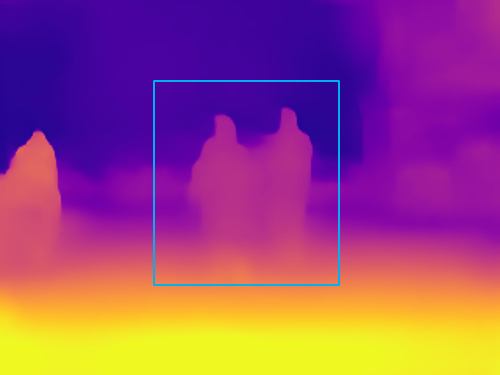} &
\includegraphics[width=0.3\linewidth]{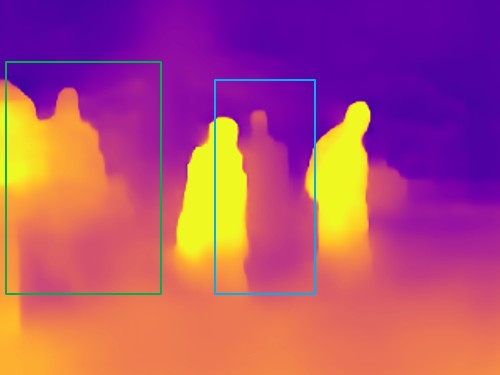} 
\\

\end{tabular}

\caption{Depth prediction results on the challenging pedestrian tracking dataset, notable improvements can be seen from our motion model.}
\label{fig:tracking}
\end{figure}

\subsection{Implementation Details}
\label{sec:implement}

Our method is implemented in TensorFlow. The input images are resized to $128 \times 416$ when training on KITTI and Cityscapes datasets, and to $240 \times 320$ when training on the pedestrian tracking dataset. The loss weights $\mu, \alpha$ are set to $ 0.5, 0.15$, respectively. The smoothness weight $\lambda$ is set to 0.04 when training on KITTI and the pedestrian tracking dataset, and to 0.008 when training on Cityscapes dataset. The $\delta$ in Eq.~\ref{eq:prior} is set to 0.01 when training on KITTI and the pedestrian tracking dataset, and to 0.5 when training on Cityscapes dataset due to the different data distributions of the datasets. The batch size is chosen to be 4 and Adam \cite{kingma2014adam} is used to optimize the network with $\beta_1=0.9, \beta_2 = 0.99$. When training the baseline model, the learning rate is set to $2 \times 10^{-4}$. Our motion model is trained with learning rate of $2 \times 10^{-5}$ and the network weights are initialized from our trained baseline model. We use the same DepthNet and PoseNet architectures as \cite{casser2019struct2depth}. Our RDN uses the same architecture as PoseNet except for the last output layer.

\textbf{Evaluation Metrics.}
We evaluate our method with  standard metrics reported in \cite{eigen2014depth}, which include 4 error metrics (Abs Rel, Sq Rel, RMSE, RMSE log) and 3 accuracy metrics ($\delta < 1.25$, $\delta < 1.25^2$, $\delta < 1.25^3$). The definition of these metrics is referred to \cite{eigen2014depth}.


\subsection{Results}

{\textbf{The KITTI and Cityscapes Datasets.}} We report our depth estimation results on standard Eigen test split \cite{eigen2014depth} of KITTI raw dataset in Tab.~\ref{tab:eigen}. All methods are evaluated on KITTI dataset, no matter whether they are trained on KITTI or Cityscapes dataset. We achieve comparable results with  \cite{casser2019struct2depth} when training on KITTI dataset, which is specially designed for outdoor driving scenes. When training on more dynamic dataset Cityscapes, our performance is consistently better than that of \cite{casser2019struct2depth}. More evidence can be seen from the qualitative comparisons in Fig.~\ref{fig:qualitative}. On the other hand, our motion model is consistently better the baseline no matter training on KITTI or Cityscapes dataset (see Tab.~\ref{tab:eigen} and the visual results in Fig.~\ref{fig:compare}).

\textbf{Ablation Study.}
To further evaluate our proposed bicubic motion representation, we create a new baseline named `Ours (Rigid)', where we replace the bicubic function in Eq.~\ref{bicubic} with rigid motion representation and keep all the other settings the same. The results given in Tab.~\ref{tab:ablation} clearly demonstrate the superiority of our bicubic motion representation, winning the majority of the metrics.

To evaluate the contribution from the object depth prior, we add another baseline called `Ours (w/o prior)', which disables the prior in our full motion model. As shown in Tab.~\ref{tab:ablation}, compared with Ours (Motion), the performance of Ours (w/o prior) degrades a lot, which verifies that our object prior is simple yet efficient to handle the infinite depth issue as explained in Sec.~\ref{sec:prior}.


\textbf{The Pedestrian Tracking Dataset.}
To illustrate the generality of our proposed bicubic motion representation, we conduct experiments on a crowded pedestrian tracking dataset, which 
is quite different from KITTI and Cityscapes datasets and particularly challenging due to the presence of many deformable pedestrians. Fig.~\ref{fig:tracking} visualizes the depth prediction results of samples from this dataset. Clear improvements can be seen from our motion model. The promising results show the generality of our proposed bicubic motion representation and indicate that  our framework is applicable to unconstrained monocular videos.

\section{Conclusion}

We have presented a learning based approach to estimate depth from unconstrained monocular videos without ground truth supervision. The approach consists of DepthNet, PoseNet and RDN, for which a deformation based bicubic motion representation is proposed to model object motions in diverse scenes. The experimental results on several datasets show the promising performance of the proposed approach and validate the effectiveness of the deformation based motion representation as well. 


\section{Acknowledgements}

The authors are supported by the National Natural Science Foundation of China (No.~61672481) and the Youth Innovation Promotion Association CAS (No.~2018495).

\bibliographystyle{named}
\bibliography{ijcai19}

\end{document}